# Deep learning achieves radiologist-level performance at segmenting breast cancers on MRI

Lukas Hirsch, MS[1]*, Yu Huang, PhD[1,2]*, Shaojun Luo, PhD[6], Carolina Rossi Saccarelli, MD[2], Roberto Lo Gullo, MD[2], Isaac Daimiel Naranjo, MD[2], Almir G.V. Bitencourt, PhD[2,7], Natsuko Onishi, MD, PhD[2,3], Eun Sook Ko, PhD[2,4], Doris Leithner, MD[2], Daly Avendano, MD[2,5], Sarah Eskreis-Winkler, MD, PhD[2], Mary Hughes, MD[2], Danny F. Martinez, MS[2], Katja Pinker, MD, PhD[2], Krishna Juluru, MD[2], Amin E. El-Rowmeim, MS[2], MA[2], Pierre Elnajjar, MS[2], Elizabeth A. Morris, MD[2], Hernan A. Makse, PhD[6], Lucas C Parra, PhD[1+], Elizabeth J. Sutton, MD[2]

1 Department of Biomedical Engineering, the City College of the City University of New York, New York, NY 10031

2 Department of Radiology, Memorial Sloan Kettering Cancer Center, New York, NY 10065

3 Department of Radiology, University of California, San Francisco, CA 94115

4 Department of Radiology, Samsung Medical Center, Sungkyunkwan University School of Medicine, Seoul, Korea

5 Department of Breast Imaging, Breast Cancer Center TecSalud, ITESM Monterrey, Monterrey, Nuevo Leon, Mexico.

6 Levich Institute and Department of Physics, the City College of the City University of New York, New York, NY 10031

7 Department of Imaging, A.C.Camargo Cancer Center, São Paulo, Brazil

* first authors contributed equally
+ corresponding author

**Funding Information:** Support for this work was provided in part by the National Institute of Health (NIH), National Science Foundation (NSF) through the following grants: NIH P30 CA008748; NIH R01EB022720, NIH R01MH111896; NIH R01EB022720, NSF DRL-1660548, NIH-NCI R01CA247910, as well as from the Breast Cancer Research Foundation and the Spanish Foundation Alfonso Martin Escudero.

**Originating Institution:** Memorial Sloan Kettering Cancer Center, 1275 York Avenue, New York, NY 10065

**Corresponding Author**: Lucas C Parra, Ph.D, The City College of New York, 160 Convent Avenue, New York, NY 10031

**Senior Author**: Elizabeth Sutton, MD, Memorial Sloan Kettering Cancer Center, 1275 York Avenue, New York, NY 10065

## Abstract

**Purpose:** The purpose was to develop a deep network architecture that achieves fully-automated radiologist-level segmentation of cancers in breast MRI.

**Materials and Methods:** We leveraged 38,229 exams (64,063 individual breast scans) collected retrospectively from women aged 12-94 (mean age 54) who presented between 2002 and 2014 at a single clinical site. For network training, we selected 2,555 breast cancers which were segmented in 2D by radiologists, as well as 60,108 benign breasts, which served as examples of non-cancerous tissue during training. For testing, an additional 250 breast cancers were segmented independently in 2D by four radiologists. We selected among several 3D deep convolutional neural network architectures, input modalities and harmonization methods. The outcome measure was the Dice score for 2D segmentation, and was compared between the network and radiologists using the Wilcoxon signed-rank test and the TOST procedure.

**Results:** The best-performing network on the training set was a 3D U-Net with contrast enhancement dynamic as input and with intensity normalized for each exam. In the test set the median Dice score of this network was 0.77+/-0.26. The performance of the network was equivalent to that of the radiologists (TOST procedure with radiologist performance of 0.69-0.84 as equivalence bounds: p < .001 and p < .001, respectively; N = 250) and compares favorably with published state of the art (0.6-0.77).

**Conclusion:** When trained on a sufficiently large dataset, a carefully designed 3D U-Net performs as well as fellowship-trained radiologists in detailed 2D segmentation of breast cancers in routine clinical MRI.

## Introduction

Segmentation of breast tumors provides image features such as shape, morphology, texture, and enhancement dynamics that can improve diagnosis and prognosis in breast cancer patients (1–3). Reliable automated tumor segmentation does not yet exist, and manual segmentation is labor-intensive; this has precluded routine clinical evaluation of tumor volume despite mounting evidence that it is a good predictor of patient survival (2). Automatic segmentation with modern deep network techniques has the potential to meet this clinical need. Deep learning methods have been applied in breast tumor segmentation (4,5) and diagnosis (6–11) in mammograms, where large datasets of up to one million images are available, which greatly boosts the performance of the machine-learning systems (12,13). Yet unlike MRI, mammograms cannot determine the exact 3D location and volumetric extent of a lesion. Breast MRI has higher diagnostic accuracy than mammography (14–16) and outperforms mammograms in detecting residual tumors after neoadjuvant therapy (17). Additionally, background parenchymal enhancement (BPE) measured in MRI with dynamic contrast enhancement (DCE) is predictive of cancer risk (18). Several studies have automated tumor segmentation in breast MRI using modern deep networks such as U-Nets or DeepMedic (19–25), focusing mostly on malignant tumors. With reliable fully-automated segmentation, the overall clinical workflow could be improved, as such segmentation could aid radiologists in tumor detection and diagnosis. In research, fast automated segmentation might help identify important prognostic and predictive biomarkers. Unfortunately, the available MRI datasets to train segmentation algorithms have remained comparatively small with 50–250 MRI exams (19–25), thus limiting the potential of modern deep networks. Studies are sometimes limited to semi-automated segmentation (26) and performance differs across datasets making comparison to radiologists difficult. Where a formal comparison was conducted on the same dataset, radiologists still outperform the networks (20). We hypothesize that human-level performance can be achieved if a sufficiently large dataset is used to train a modern deep convolutional network. The goal of this research was to develop a deep-network architecture that achieves fully-automated radiologist-level segmentation of breast cancer on MRI by leveraging a dataset of over 60 thousand breast MRIs.

## Materials and Methods

### Study Design

This study was approved by the Institutional Review Board and written informed consent was waived due to the retrospective nature of this study. All data handling complied with HIPAA regulations.

The retrospective dataset was composed of 38,229 clinical MRI breast exams performed from 2002–2014 for either high-risk screening (11,929 women) or diagnostic purposes (2,546 women). The age range of the patient population was 12–94 years (mean age 52 years). Exams included unilateral and bilateral examinations totaling 64,063 breasts (see Fig. 1). Of these, 3,955 breasts had biopsy-confirmed malignant pathology (malignant breast), and 60,108 had benign pathology or showed 2 years of imaging stability (BI-RADS ≤ 3) and/or no clinical evidence of disease (benign breasts). The types of tumors included in the study are listed in Table S2.

The data were partitioned into training and test data at random (see Fig. 1, Table S1). For the purpose of training the segmentation network, voxels were labeled as tumor-positive or negative using 2D segmentations performed by fellowship-trained radiologists on malignant breasts (see details below). Training of the network also included voxels of non-cancerous tissue taken from a central MRI slice of benign breasts, which served as additional negative control examples. Therefore, the network was trained to distinguish cancerous from non-cancerous tissue. Using the training data, we first selected the best performing network architecture, established the utility of the available imaging sequences, and selected an effective harmonization procedure.

We then compared the segmentations produced by the final network to those produced by fellowship-trained radiologists in a separate test set of 250 malignant breasts which were each segmented independently by four radiologists in 2D (see below). Benign breasts were not included in this analysis. The goal of this analysis was to determine if the performance of the network is equivalent to the performance of fellowship-trained radiologists at segmenting pathology-confirmed cancers.

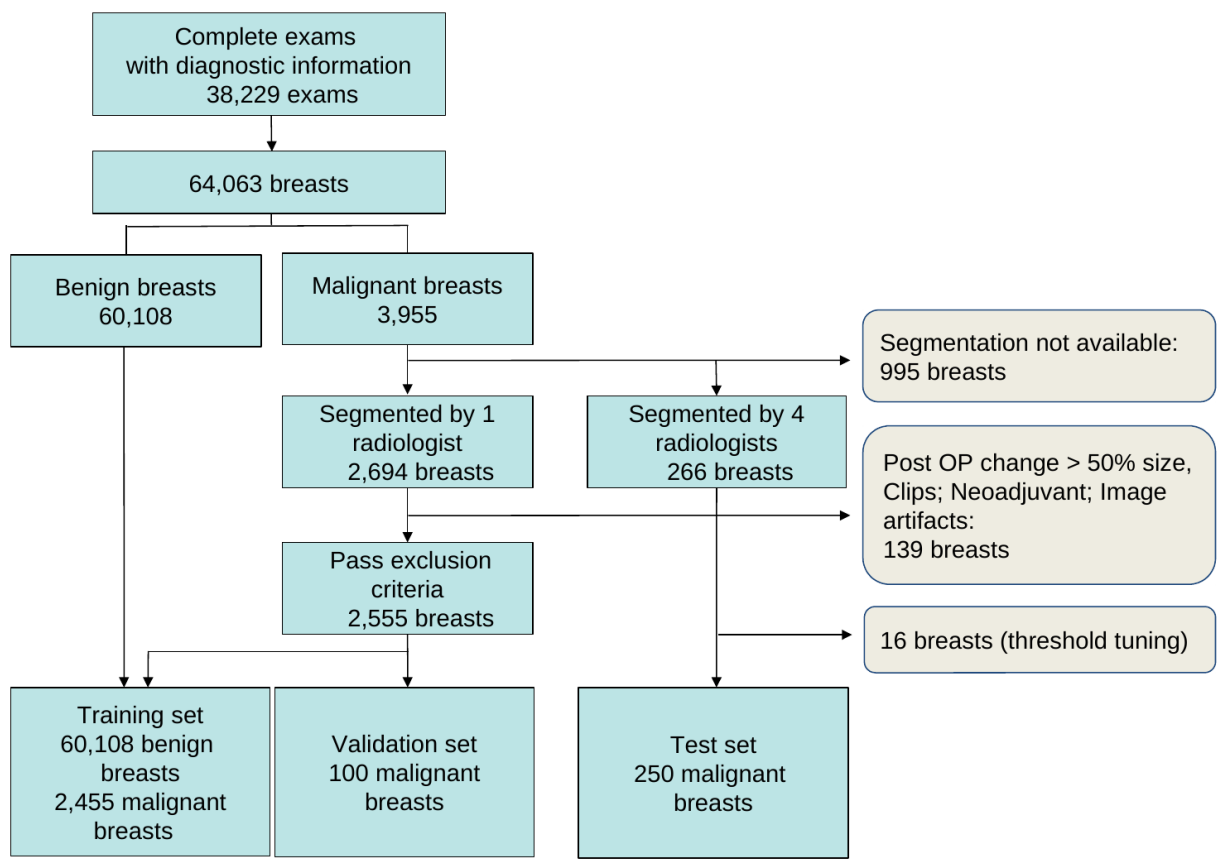

**Figure 1: Number of exams and breasts used in training and testing.** See also Table S1.

## Data Description and Harmonization

All breast MRI examinations were acquired on either a 1.5 or 3.0 Tesla GE scanner (GE Medical Systems, Waukesha, WI). Exams were acquired in the sagittal plane (Fig. 2A) at varying in-plane resolutions (Fig. 2B), 2–4 mm slice thickness, and varying repetition times and echo times. The sequences used from each MRI examination included pre-contrast, fat-saturated T2- and T1-weighted images, and a variable number (N = 3–8) of post-contrast fat-saturated T1-weighted images (T1c). In-plane sagittal resolution was harmonized by upsampling relatively low resolution images by a factor of two (Fig. 2B). Image intensity from different scanners were harmonized by dividing with the 95th percentile of pre-contrast T1 intensity. To summarize the DCE we measured the volume transfer constant for the initial uptake and subsequent washout, DCE-in and DCE-out respectively (Fig 2A; DCE-in is the first post-contrast minus pre-contrast image; DCE-out is the linear slope of intensity over time in the post-contrast images). Data collection, preprocessing and harmonization are described in more detail in the Supplement.

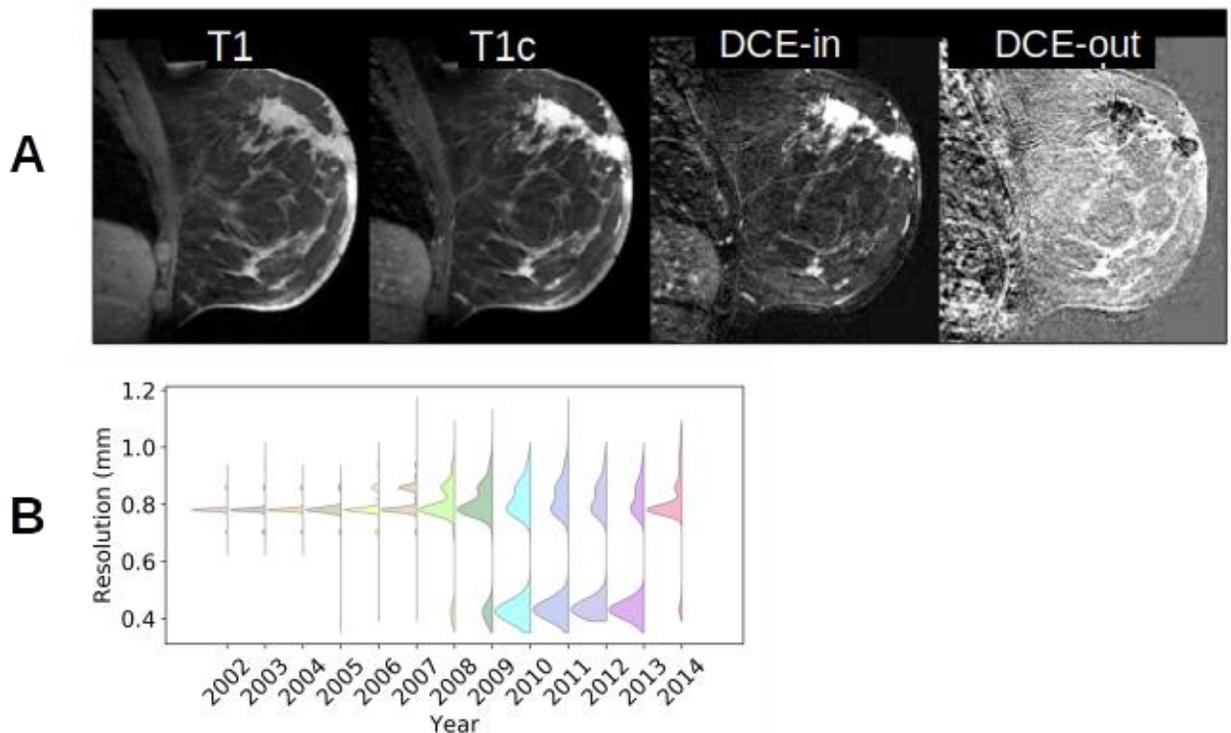

**Figure 2. A:** Example of pre- and first post-contrast fat-saturated images, T1 and T1c, respectively. Initial dynamic contrast enhancement in this malignant breast is evident after subtraction of the aligned T1c and T1 images (DCE-in). Subsequent washout (DCE-out) is evident in the subsequent drop of intensity, measured as slope over time. **B:** T1c scans accrue between 2002-2014 with a range of in-plane resolutions.

### Radiologist Segmentations

All segmentations were performed in 2D slices by fellowship-trained breast radiologists (R1–R10, years of experience: 5, 8, 5, 5, 8, 12, 7, 13, 5, 2, i.e. fellowship training and subsequent clinical practice). To train the network, at the start of this project 2,694 breasts had been segmented by individual radiologists (R1–R10) by outlining the malignant tissue in a single slice. These segmentations were subsequently reviewed by R1-R5 to ensure they met the inclusion criteria, resulting in 2,555 segmentations used for training (Fig. 1 and Table S1; 2,455 used for training and 100 for validation). An additional 266 breast cancers which were independently segmented by all four radiologists (R1–R4); of these, 16 were used for threshold tuning and 250 for testing. The common 2D slice to be segmented was selected by R5, containing the largest area of the index cancer. Radiologists performed segmentations on the T1c image, with the T1 and fat-saturated T2-weighted images available for reference. For the test data, we also provided the T1c-T1 subtraction (DCE-in) which quantifies initial uptake. See the Supplement for more details.

### Convolutional Neural Networks

We used networks based on the DeepMedic network (27) and a 3D U-Net (20), which have been used extensively for medical segmentation, including breast segmentation (19–24). The architecture of the 3D U-Net is described in Fig. 3 (Fig. S2 for the DeepMedic). Following previous studies (28) the traditional space-invariant implementation has been augmented by adding a spatial prior as input to the final classification. For the U-Net, the spatial prior was a breast mask, as in previous studies (20). This breast mask was computed using a separate U-Net operating on the entire image at lower resolution. This network was trained on a smaller number of manually segmented breast slices (N = 100; performed by the first author). To avoid blocking artifacts that are often observed in U-Nets (29), we carefully re-designed the conventional downsampling and upsampling steps. The 3D U-Net had approximately three million parameters. Details of the architectures, sampling, and training of the network parameters are in the Supplement (Fig. S2, S3).

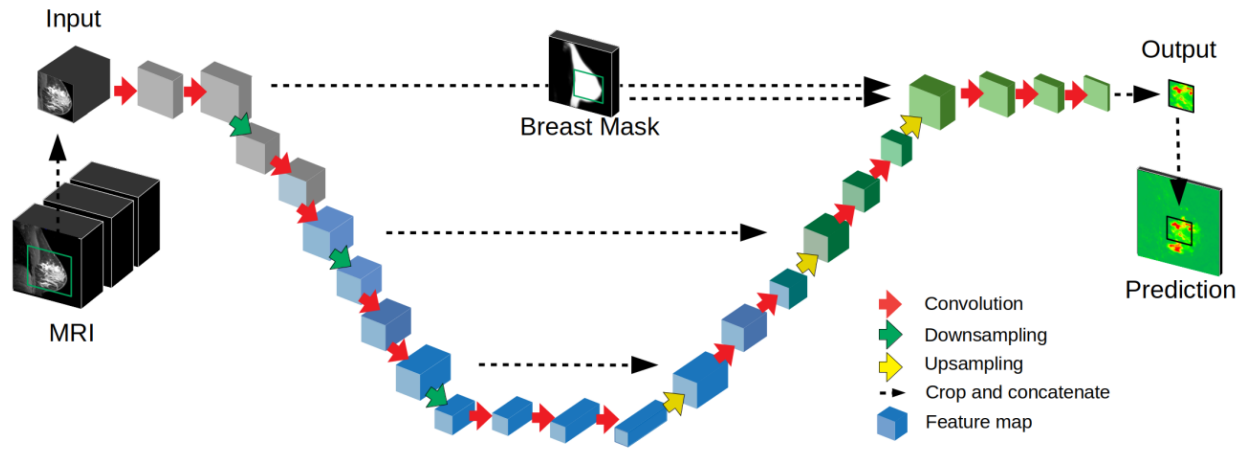

**Figure 3: Deep convolutional network used for segmentation.** We used a 3D U-Net with a total of 16 convolutional layers (red arrows) resulting in 3D feature maps (blue blocks). The input MRI includes several modalities (see Fig. 2A). The network output is a prediction for a 2D sagittal slice, with probabilities for cancer for each voxel (green/red map). The full volume is processed in non-overlapping image patches (green square on 'Input MRI'). A breast mask provides a spatial prior as input to the U-Net.

## Primary Outcome Measure

The network estimates for each voxel the probability of being part of the cancerous tissue (Fig. 4A - M Probs). A binary segmentation was obtained by thresholding this probability at a fraction of the maximum in the selected slice (Fig. 4B - M) and dismissing disconnected areas that did not reach the maximum. The primary outcome measure was the Dice score (30) for the cancer evaluated on the 2D slice, with consensus segmentation as the reference (e.g. Fig. 4A - Ref4; and Supplement for detail). A Dice score of 1.0 corresponds to perfect overlap and a score of 0.0 indicates no overlap. To determine the sample size required for the comparison between network and radiologists we assumed a mean Dice score of 0.75 as lower bound for the radiologist and 0.80 for the network. Power analysis for a paired t-test was performed with logit-transformed Dice scores, to approximate normality (31). An estimate of standard deviation of 1.5 across exams (based on the network performance in the validation set) resulted in a Cohen's d of 0.1918. With this effect size a power of 85% at a significance of 0.05 requires 246 cases. We selected a sample size of N = 250 for the test set.

## Statistical Methods

All pairwise performance comparisons among different network architectures and between the network and radiologists used the Wilcoxon signed-rank test on the Dice score. This accounts for the deviation from normality observed for the Dice scores (Shapiro–Wilk test, W = 0.61, p < .001 for difference in mean Dice score, as well as for logit-transformed Dice scores W = 0.92, p < .001). We tested for equivalence in the Dice score of human and machine using the TOST procedure (32). In this procedure performance is compared to a bottom and top equivalence bound, for which we selected the bottom and top performing radiologists. The Wilcoxon signed-rank test was used instead of the conventional t-test of the TOST procedure because Dice scores were not normally distributed. Effect size (ES) of the Wilcoxon test is reported following (33). Medians are accompanied with the interquartile range written as +/-IQR.

# Results:

To select the preferred network architecture, input modalities, and harmonization method we trained various architectures (as described in the Supplement). We evaluated performance on a validation set, i.e. a fraction of the training data that is left out from training. Based on these results (Fig. S4) we selected a volumetric implementation of a 3D U-Net (Fig 3), which takes as inputs the first post contrast T1 (T1c), the initial uptake (T1c – T1), and the slope of the subsequent contrast (T1c-slope) (Fig. 2A). These sagittal images were aligned with deformable co-registration (34) covering the volume of a single breast. Intensity was harmonized by scaling each exam separately with a robust maximum of the pre-contrast T1. When training this network on a dataset of different sizes we confirm the main hypothesis that the larger dataset significantly improves Dice score performance (0.63, 0.69 and 0.73 when training on 240/240, 2,400/2,400 and 2,455/60,108 malignant/benign breasts, respectively, evaluated on a separate validation set of 100 malignant breasts; Fig. S5).

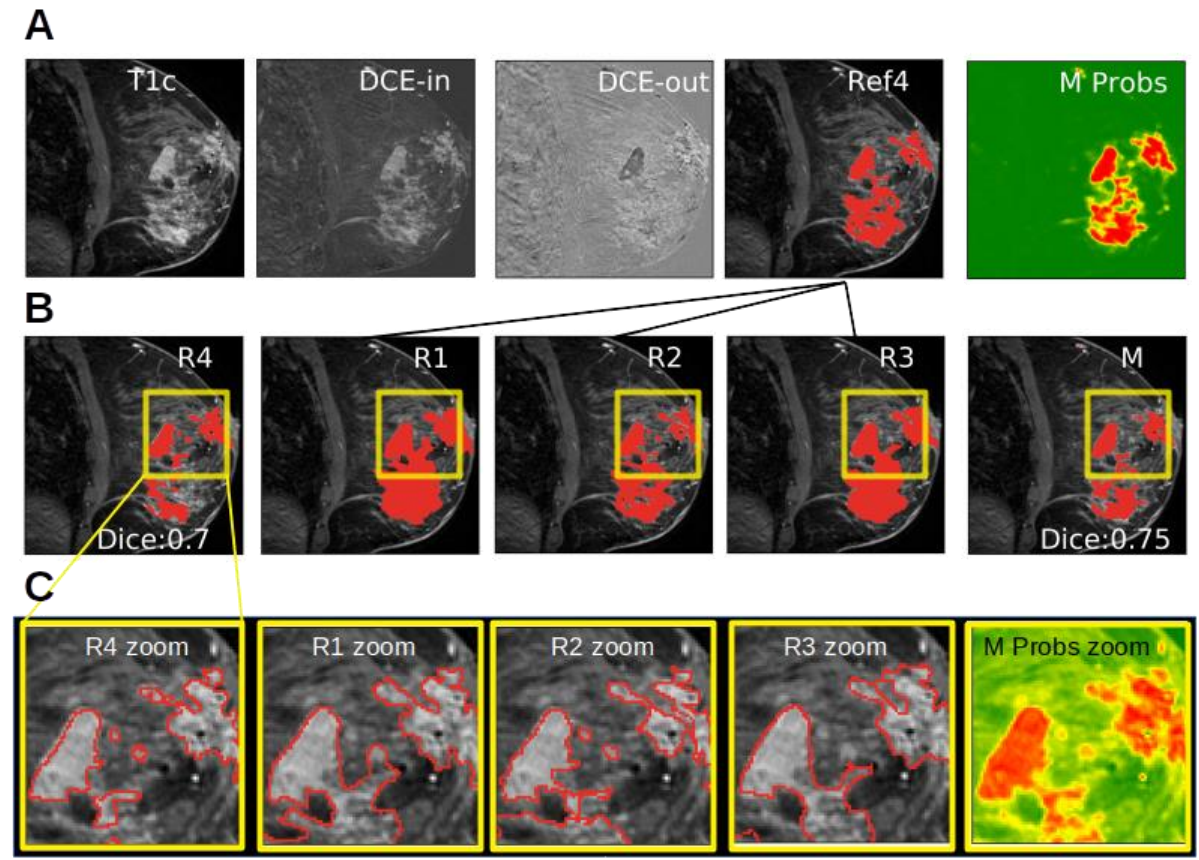

**Figure 4: Manual and automated segmentations of breast cancer. A:** Inputs to the model consisting in first post-contrast (T1c), T1c-T1 subtraction (DCE-in) and washout (DCE-out), with an independent reference for R4 made from the intersection of R1–R3 (Ref4) and the output of the network (M Probs) indicating probability that a voxel is cancer (green = low; red = high). **B:** Example segmentation of all four radiologists (R1-R4) for a given slice, and the model segmentation created by thresholding probabilities (M). Dice scores for R4 and M are computed using Ref4 as the target. **C:** Zoom in for the area outlined in yellow. Here showing the boundaries of segmentations for the machine as well as human segmentations as drawn on the screen by radiologists R1-R4.

Performance of this final design was tested on an independent test of 250 malignant cases. While the network produced 3D segmentations, evaluation was limited to 2D segmentations, with cancers segmented independently by four breast radiologists (R1–R4) in a single 2D slice per breast (Fig. 4). Segmentations differed across radiologists in the areas selected and detailed boundaries (Fig. 4C, and Fig. S9). An independent reference segmentation was obtained for each radiologist using segmentations from the three remaining radiologists (e.g Ref1 is the intersection of the segmentations of R2–R4 and is used to evaluate R1; Fig. 4A - Ref4). The threshold for converting continuous probabilities at the output of the network into binary segmentations was estimated using a separate set of reference segmentations (N = 16 not included in the test set, Fig. 1, Fig. S6A). The resulting Dice score (averaged over the four references) had a $5^{th}$–$95^{th}$ percentile range of 0.43–0.90 for the radiologists and 0.21–0.92 for the network (Fig. 5A). These mean Dice scores did not differ significantly between the network and the radiologists (medians 0.77+/-0.26 and 0.79+/-0.15, respectively; ES = 0.51; p = 0.72, N = 250). The median Dice scores of the network were 0.76+/-0.26, 0.76+/-0.26, 0.77+/-0.28, and 0.76+/-0.28, and for the radiologists the median Dice scores were 0.69+/-0.2, 0.84+/-0.14, 0.78+/-0.13, and 0.84+/-0.13, with one value for each of the four references, indicating that the machine may have outperformed some radiologists but not others (Fig. 5B). A similar result is obtained with repeated measures ANOVA (see Supplement). To test for equivalence, we performed a TOST procedure (32) with radiologists performance as the lower and upper bounds for equivalence (R1 and R4, respectively; Fig. 5B). The Dice score of the network is significantly higher than that of R1 (ES = 0.37; p < .001, N = 250) and lower than that of R4 (ES = 0.62; p < .001, N = 250). In total, the mean performances of the network and the radiologists were indistinguishable, and median performance of the network was equivalent to that of the radiologists.

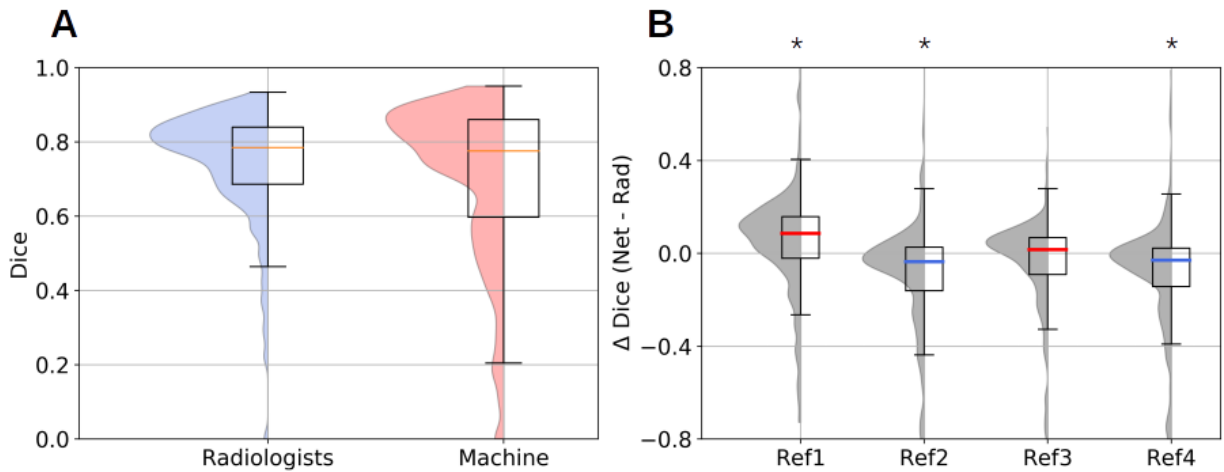

**Figure 5: Network and radiologist performance on test set. A:** Distribution of Dice scores in 250 test cases averaged across four reference segmentations. **B:** Difference in Dice score between the network and each radiologist (Δ Dice) for each of the four reference segmentations. The median Dice is higher for the network for Ref1 and Ref3 (red median Δ Dice) and higher for the radiologist for Ref2 and Ref4 (blue median Δ Dice). (*: Wilcoxon p < .001). Boxplots indicate median (in color), quartiles (box) and 1.5 interquartile range (whiskers).

There are several exams in which the network outperforms the average of the four radiologists (Δ Dice > 0, Fig. 5B; see Fig. 4 and S9 for examples). But in several instances, the network performs worse than the radiologists (Δ Dice < 0, Fig. 5B; see Fig. 6 for examples). The network deviates from the reference either in terms of the areas that it selected (Fig. 6A) or the exact boundary of the cancer (Fig. 6B). Network performance differed among tumor types (Fig. S7A), was somewhat worse in the presence of prominent BPE (Fig. S7B) and smaller cancer (Fig. S7C). Generally, images that were easy for the network were also easy for radiologists, regardless of size or BPE (Fig. S8).

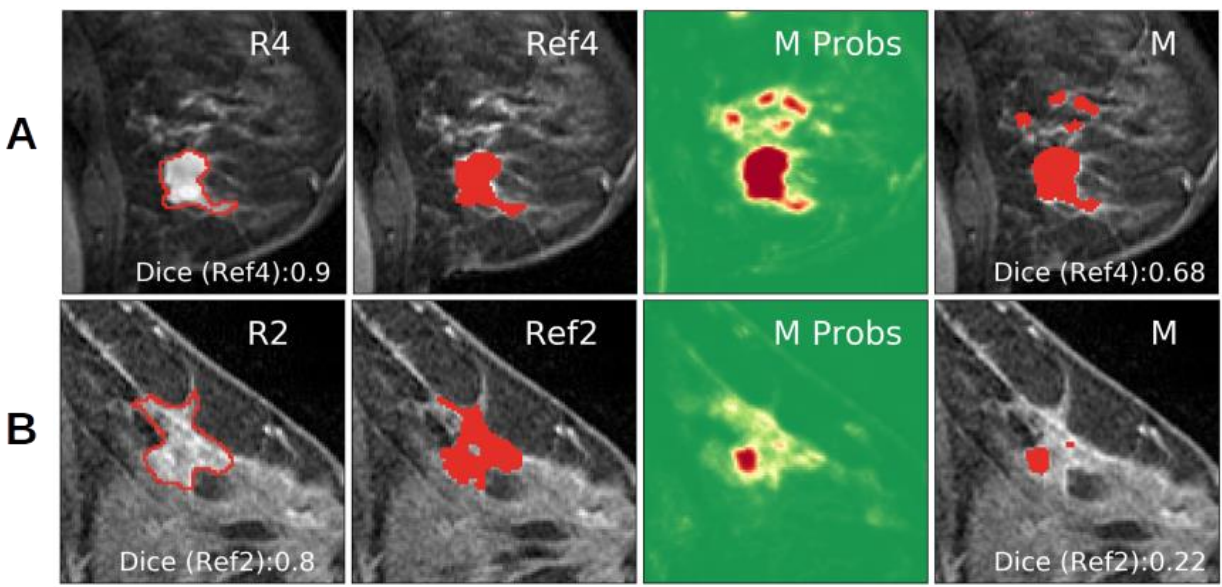

**Figure 6: Examples of cases in which the network deviates from the segmentation of the reference radiologist. A:** The network captures additional areas not selected by the radiologist R4. Dice score here shown for Ref4 (intersection of R1–R3). **B:** The network output (M Probs) captures the correct area, but after thresholding, low probability values yield a smaller region compared to the consensus segmentation (Ref2).

Similar results were obtained when using the union or majority vote of three radiologists as consensus reference (Fig. S6C). R2 and R4 outperformed the network with the intersection as consensus reference (Fig. 5B), whereas R1 and R3 outperformed the network when compared on the union as consensus (Fig. S6B). This suggests that R2 and R4 opted from more specific tumor segmentation, whereas R1 and R3 provided more sensitive segmentations (examples shown in Figure S9). As a point of reference, we also evaluated a conventional image segmentation method (fuzzy-c-means (35)) and found poor performance on these data attesting to the difficulty of the task (Dice score of 0.11, Fig. S10, and 0.65 when restricting to a limited region of interest - as in (26) - in which case the network achieves 0.82).

# Discussion

We demonstrated for the first time radiologist-level performance in fully-automated segmentation of breast cancer on MRI using routine pre- and post-contrast T1 images. The architecture of deep convolutional networks was optimized using a large training dataset of 38,000 exams, which consisted of 2,555 malignant and 60,108 benign breast scans. This dataset is substantially larger than previous efforts with deep networks, which used only 50–250 MRI exams (19–24).

Notably, the best performance was obtained with a volumetric U-Net that is conceptually simpler than previous networks (20,21,23). Complex structures with fewer parameters may have been necessary to compensate for the smaller data-set sizes used in these earlier studies. For instance, Zhang et al. combined three different 2D U-Nets in a hierarchical manner (20); one network generated a breast mask, the second produced segmentations of tumors, and the third refined these segmentations. We used a similar approach at first, following the MultiPrior network (28) with a breast mask as a spatial prior, and a conditional random field (CRF) for post-processing. However, we found that a simpler 3D U-Net without these additional modules sufficed. The U-Net also outperformed the DeepMedic (27), which we believe is the result of better integration of information at multiple spatial scales. Chen et al. used 2D U-Nets with a long-short-term-memory (LSTM) network at the input to process the contrast dynamic (21). Instead, we summarized the DCE in two images, capturing the initial contrast uptake and subsequent wash-out. This allows us to harmonize the differing sampling intervals and the number of post-contrast images, and allows the network to potentially capture something akin to the signal enhancement ratio, which has been proposed as a threshold criterion for DCE in breast MRI (36). Adoui et al. uses a Seg-Net which communicates only location information through residual connections (23). This reduces the number of parameters as compared to a U-Net, and therefore may have required fewer training images. Other previous network methods cope with smaller training sets using pre-selected features (19), unsupervised clustering algorithm (24), or by leveraging shape priors (22).

Our final U-Net implementation differs from previous approaches in two important ways. First, we used a full 3D network instead of a conventional 2D network that processes individual slices independently (12,21,23–25). While this increased the number of network parameters, it also captured volumetric features missed in 2D processing. Our implementation also avoids sampling artifacts encountered in conventional implementations of U-Nets (29). While we did this with an apparent increase in complexity, our approach obviates the need of less carefully designed architectures to "unlearn" sampling errors.

Previous efforts to apply machine learning to breast cancer segmentation report Dice scores in the range of 0.60–0.77 (19–25), but the performance of these models has generally not been compared to that of radiologists. In a head-to-head comparison, Zhang et al. (20) reported a Dice score of $0.72\pm0.28$ for the network and $0.78\pm0.30$ for the radiologists. Dice of 0.7 are already considered good agreement (31). In the present study, automated segmentations matched the detailed segmentations of a radiologist with a Dice score of 0.76-0.77. Radiologist performance here was in the range of 0.69-0.84, which is comparable to these previous reports. Equivalence of network performance was demonstrated here on a fixed test set, which is important given the variability observed across studies using different datasets (19–25). The test set here included difficult cases with small subcentimeter cancers, multicentric cancer and includes patients with breast implants, all of which have often been excluded from previous studies. Note that network performance was better on the test set as compared to the validation set (0.77 vs 0.73), which could have resulted from more careful manual segmentations when radiologists were also being evaluated. Regardless, the radiologist's performance as well as the limited performance of conventional semi-automatic image segmentation methods point to the difficulty of segmenting an entire 2D slice in detail on these diverse images. Our study is retrospective and limited to a single institution. However, the data set is heterogeneous having been collected over 12 years from different scanner types, magnet strengths (1.5 and 3.0 Tesla), different breast coils, which results in variable spatial and temporal resolutions. All these factors together add to the difficulty and clinical realism of this study.

The network classifies each voxel in the image and thus in principle provides volumetric segmentation. The main clinical utility is to facilitate volumetric assessment, which is not broadly used despite proven benefit (2,3). However, we evaluated automated segmentations only in 2D slices because we expected better radiologist performance as compared to highly labor-intensive 3D manual segmentation. Another limitation of the work is that we used sagittal images, which was standard practice at our institution until 2014, and which therefore characterized most of the historical data. Breast MRI protocols are often acquired in the axial plane and with continued improvements in technology, with higher temporal and spatial resolution. Future studies could focus on volumetric evaluation of segmentations. Additionally, for high-resolution multi-planar breast MRI one might expect to achieve higher segmentation performance for both the network and radiologists. To facilitate such studies we make all code and the pretrained network freely available on github.

# Supplementary Material

| MRI exams | 38,229 |
|---|---|
| Scanned breasts | 64,063 |
| Breasts with biopsy-proven cancer | 3,955 |
| Benign breasts used for training | 60,108 |
| Malignant breasts used for training | 2,555 segmented by 1 radiologist (2,455 training set, 100 validation set) |
| Malignant breasts for used threshold tuning | 16 segmented by 4 radiologists |
| Malignant breasts used for testing | 250 segmented by 4 radiologists |

**Table S1: Summary of available data used in training and testing.** Note that most patients had multiple screening breast MRIs. Thus, the dataset contains multiple scans of the same breast at different time points. The partition into training/validation/testing sets did not include the same breasts in different sets.

**Data Description and Preprocessing**

All breast MRI examination data were acquired with GE MRI scanners (GE Medical Systems, Waukesha, WI) with either a 1.5 or 3.0 Tesla field and dedicated 4, 7, 8, or 16-channel breast coils. Five different scanner types were used: Signa Excite, Genesis Signa, Discovery, Signa HDxt and Optima MR450w. Exams were acquired in the sagittal plane (Fig. 2A) at various in-plane resolutions (Fig. 2B), 2–4 mm lateral resolutions, and varying TR and TE times. Intravenous gadolinium-based contrast agent was administered at 0.10 mmol per kg of body weight and at a rate of 2 ml/sec. Prior to analysis, data were de-identified by removing all patient information and by saving exams with anonymized patient identifiers. DICOM headers were used to identify the modality (T1 or T2), laterality (left or right breast), fat saturation, and acquisition time for pre- and post-contrast T1 images. In total, data was available for 38,229 examinations with complete imaging, BIRADS, pathology and clinical follow-up information for 64,063 breasts (some examinations were unilateral).

**Harmonization**

To compensate for the diversity of data acquisition techniques in the 12-year period in which the study data was collected, it was necessary to harmonize the spatial resolution, intensity, and contrast dynamics of the data. In-plane sagittal resolution was harmonized by upsampling low resolution images by a factor of two to approximately match higher resolution images (using the resize function in Python package scikit-image) (Fig 2B). Lateral resolution was left unchanged. Images from different scanners were harmonized by using intensity normalization, similar to previous work (37). For this purpose, a breast mask was established for each T1 image (see below). Then, for each scanner, we computed a histogram of all breast voxels and calculated a nonlinear transformation to transform these histograms into a standard distribution (Fig. S1). We deviate from (37) by using a Chi-square of degree k = 4 as target distribution. This transformation was then applied to all image intensities of that scanner, including pre- and post-contrast images. We also used an intensity harmonization based on each exam by dividing all images from an exam with the 95th percentile of the pre-contrast T1 intensity. This simpler exam-based alternative was ultimately adopted as it outperformed the scanner-based intensity harmonization (see Fig. S4D). Finally, to harmonize the contrast dynamic, we computed the initial uptake (DCE-in) and the intensity slope over time of the post-contrast MRIs to capture DCE wash-out (DCE-out) measured in units of $\text{min}^{-1}$, i.e. volume transfer constants (Fig.2A). This initial uptake and subsequent slope were computed after deformable co-registration with NiftyReg (34), where we used the first post-contrast image T1c as a common reference for all images.

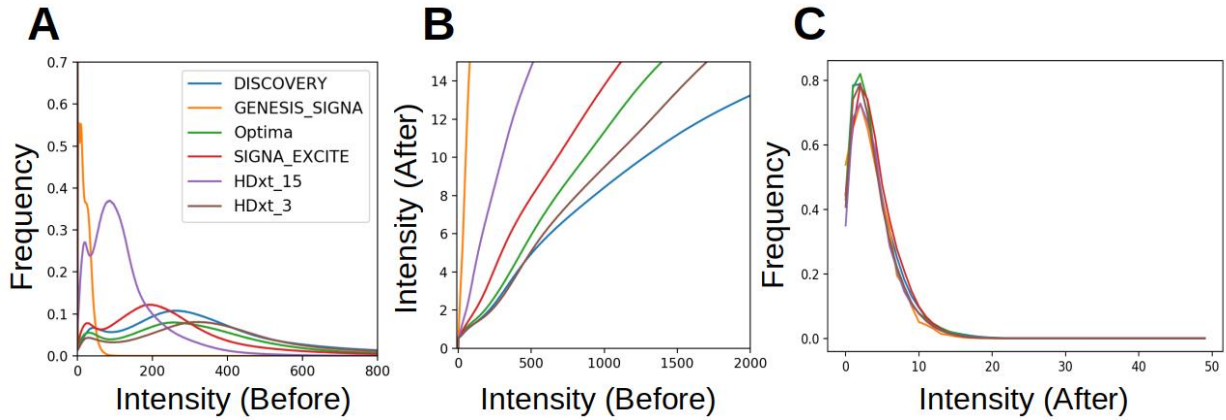

**Figure S1.** Intensity harmonization across different scanners to a standard Chi-square distribution: **A:** Intensity distribution of each scanner and B-field intensity (each distribution belongs to one of a total of six scanner manufacturers identified). **B:** Non-linear transformation mapping the intensities from each scanner to intensities of a target distribution (same colors as in panel A). **C:** Intensities distribution after the non-linear mapping transformation. They all approximate the target of a Chi-square distribution (with k = 4).

**Details on Radiologist Segmentations**

From the available 3,955 confirmed malignant exams we completed segmentations on 2,694 exams (Fig.1). These exams included all required sequences (T1, T1c, and T2), were of adequate image quality, and were free from gross imaging artifacts. Of these 2,694 exams, radiologists R5–R8 segmented an initial 880 exams, whereby segmentation was limited to the largest index cancer in a single 2D slice. Radiologists R1–R4 segmented an additional 1,814 exams, whereby segmentation included the entire extent of disease included in the 2D slice to provide the network a more complete sample of cancer voxels. Note that benign voxels for training are taken only from benign breasts, so that segmentations performed by R5-R8 on the largest index cancer can still be used for training.

All breast cancer segmentations were reviewed by radiologists R1–R5 and exams were excluded in any of the following circumstances: significant post-surgical change with partial resection of the cancer, post-biopsy change larger than 50% of the index cancer, MRI performed of breast cancer after having been treated with neoadjuvant chemotherapy, and inaccurate segmentation that was missing more than 50% of the mass. Of the 2,694 exams, 139 were excluded resulting in a total of 2,555 segmented breast cancer exams, which were used for network training.

An additional 266 malignant breasts were selected at random with the same inclusion/exclusion criteria. Each were segmented by four radiologists (R1–R4) in a single 2D slice selected by R5. These segmentations included the entire extent of the disease. Of these 16 are used for threshold tuning (see Fig. S6) and 250 an independent test set to evaluate the network and radiologists.

Radiologists performed segmentations on the first post-contrast image (T1c) with the pre-contrast T1 and T2 images available for reference, along with the existing radiology reports. For the test data, we also provided the initial uptake image (DCE-in). All segmentations were performed manually with a custom graphic user interface (based on the Tkinter Python toolbox) using a tablet pen or mouse (based on personal preference). Radiologists on average spent approximately 12 minutes segmenting one cancer slice by drawing the outline (including holes) and then automatically filling the cancer area(s).

The Dice score was used for evaluation of the segmentation generated by the network against the radiologist's segmentation. This metric is common in segmentation tasks (20,28), and is defined as: $Dice = 2TP / (2TP + FP + FN)$ where TP = true positives, FN = false negatives, and FP = false positives, corresponding to classified voxels against the defined reference.

**Implementation Detail for DeepMedic**

The DeepMedic network follows the implementation of Hirsch et al. (28) and was based on the architecture proposed in Kamnitsas et al. (27). The general architecture is shown in Fig. S2. The network consists of a 3D convolutional neural network (CNN) with 11 layers and two parallel pathways. All layers contain learnable kernels (size 3×3×3) and ReLU as nonlinear activation functions, except the last classification layer, which has a Softmax activation function. During training, this network takes as input a patch of 13×35×35 voxels around a target patch of 9×9 to be classified and a larger field-of-view of 13×75×75 voxels at half-resolution for extracting contextual features. As with the 3D U-Net, we add a TPM to provide prior spatial information. Here the TPM was the average segmentation of the cancers in the training data for all breasts after rigid body registration. Prior probabilities for the target patch are extracted from a TPM and added as input to the final classification. This TPM is constructed by averaging the segmentations from the training set. The final three layers take the concatenated output of all three pathways as input and classify the target patch with fully connected layers and no additional spatial mixing (kernel size 1×1×1). In total, the DeepMedic network had approximately one million parameters.

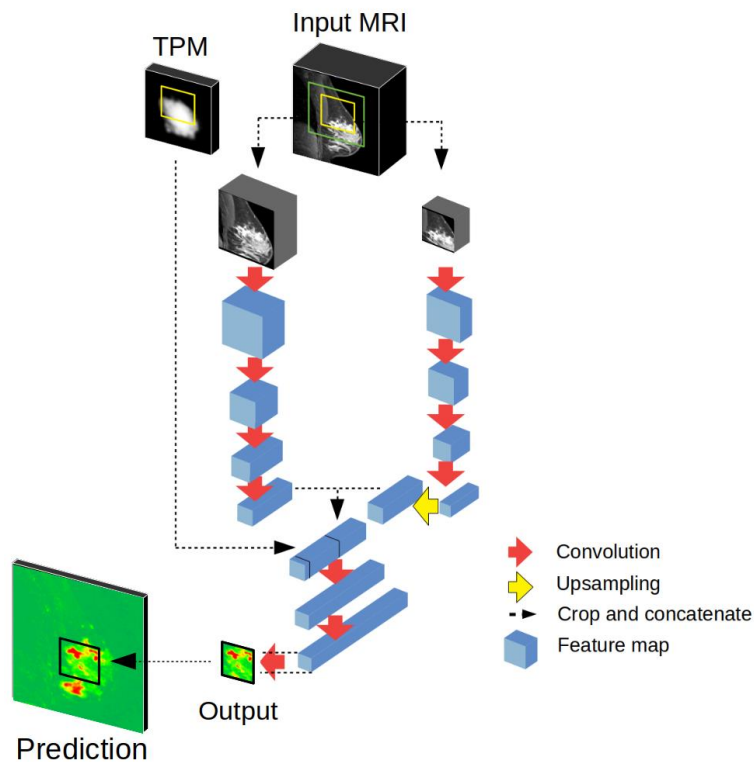

**Figure S2**: The DeepMedic network processes an image section or patch at high-resolution (contoured in yellow) and in parallel a larger patch (contoured in green) at half the original resolution. The output is a 2D image patch, with probability predictions per pixel for the center of the input image volume. A Tissue Probability Map (TPM) is added as extra features, which was made from averaging cancer segmentations from the training set.

**Implementation Detail for 3D U-Net**

The 3D U-Net also follows the implementation of Hirsch et al. (28). The U-Net is a 3D CNN with 16 layers. The input is an image patch of dimensions 19×75×75 at full resolution (see Fig. 3, green square in input), which passes through all convolutional layers with learnable kernels of size 3×3×3 and ReLU as nonlinear activation functions. The layers are organized in convolutional blocks, each consisting of two convolutional layers, one downsampling layer in the first half of the network, and upsampling layers in the second half of the network. Downsampling is performed with average pooling layers, and upsampling with transposed convolutional layers with fixed weights (see Implementation of Upsampling and Downsampling in the U-Net). Residual connections concatenate the feature maps of each block on the same scale (see Fig. 3). The last layer has kernels of size $1^3$ with a Softmax activation function, which outputs a prediction for the center 37×37 area of the middle sagittal slice of the input (see Fig. 3, black square on prediction).

Implementation and training of the DeepMedic and the U-Net were performed in Python 2.7, with Keras 2.3.1 using TensorFlow backend 1.14.0, and NumPy version 1.16.6. Initialization of the kernel weights of the DeepMedic was done with Keras's Orthogonal initializer. The U-Net's kernel weights were initialized using Keras's default glorot uniform initializer. Both NumPy and

TensorFlow were seeded with seed numbers 1 and 2, respectively, for initialization of all random numbers.

**Implementation of Upsampling and Downsampling in the U-Net**

Conventional implementations of U-Net are prone to blocking artifacts (29). We carefully designed the downsampling and upsampling steps to avoid these artifacts. In all layers we selected "valid" padding, which reduced the size of the feature maps on each convolutional layer such that the model depth was kept relatively low at 4 downsampling blocks and 4 upsampling blocks. Downsampling was performed using average pooling with a kernel of size 3 and stride 2. Upsampling used a transposed convolution with fixed kernels of size 3, with fixed weights (0.5, 1, and 0.5) to perform exact bilinear interpolation. Residual connections between downsampling and upsampling pathways required cropping borders by 1 voxel to match feature map dimensions. We note that this implementation deviates from common implementations of the U-Net. Typically, U-Nets use average pooling with variable kernel sizes for downsampling and learnable weights in the transposed convolutional layers for upsampling, but this can cause border and checkerboard artifacts (29). Fixing the transpose convolution parameter, as we have done here, seems to limit the flexibility of the network while imposing additional processing steps; however, this is worthwhile to avoid blocking artifacts. Additionally, this reduces the number of trainable parameters and provides a better training starting point by ensuring artifact-free upsampling from the start.

**Network Training**

The network is trained using conventional gradient descent to minimize a generalized Dice loss (28) using the Adam optimizer (38) (Fig. S3A and Fig. S3B). We used a batch of size 16 for each weight-updating iteration. The Dice score for the validation set was monitored at every epoch (Fig. S3C), which occurred every 2,976 weight update iterations with 48,000 image patches drawn from 4,800 scans per epoch. For data augmentation, each input was randomly rotated on one of three rotation axes. The learning rate was set to $10^{-6}$, and adapted automatically with the Adam optimizer (38). Learning was halted when we observed overtraining on the validation set, which is a subset of the training data not used for training (100 malignant scans). During training, image patches were presented containing cancerous and non-cancerous regions in a ratio of 1:4, as the benign tissue samples available far outweighed the malignant tissue samples available. Cancer voxels were sampled uniformly across all segmented slices. Non-cancerous voxels were sampled from center slices of benign breasts, enriching for locations with high T1c values. This sampling process focused the training on the areas of benign breasts that are most challenging to classify due to BPE. No strong overfitting behavior was observed, so regularization was not applied, such as dropout or L2 penalty.

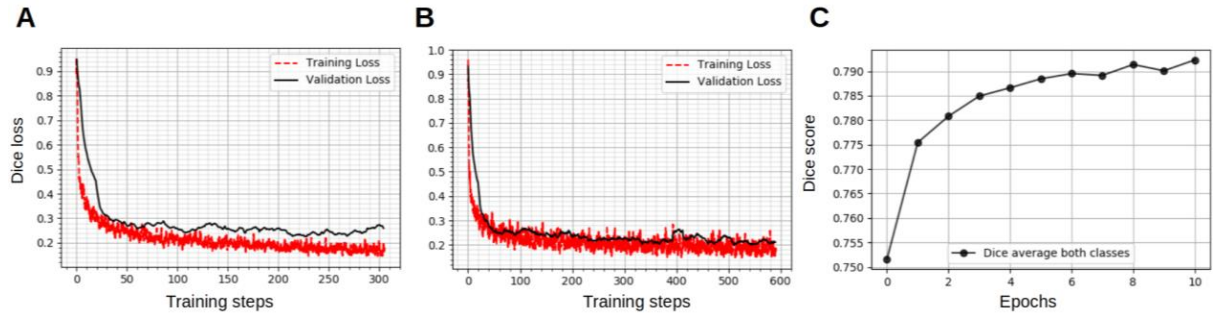

**Figure S3:** Training curves for the network. **A:** Training (red) and validation (black) loss during training on a set of 240 malignant and 240 benign breasts showing clear signs of overfitting (the loss keeps decreasing on the training data (red) but starts increasing on the validation data (black)). **B:** Training (red) and validation (black) set loss on the full training set (2,555 malignant and 60,108 benign breasts). Both losses decrease during training, showing asymptotic behavior and no overfitting. **C:** Average Dice score for both background and foreground voxels taken each epoch on the validation set containing 100 malignant scans. Each epoch consists of a full pass through all malignants in each set.

## Network Architecture Optimization

We organized the architecture optimization as a series of choices between alternative methods evaluated based on the Dice score of the validation set. In all comparisons that follow, we report the number of malignant exams used for training and validation (Nt and Nv, respectively), the median Dice score for the two methods on the validation set, and the p-value of a paired test on the difference of medians.

First, we used a MultiPrior network (28), which is a DeepMedic network (27) with additional prior information (Fig. S2). Specifically, it included a TPM to provide spatial priors on likely location of cancers and a CRF to provide morphological spatial priors. Adding this TPM improved performance by removing erroneous malignant voxels outside the breast (Nt = 869; Nv = 29; median Dice without TPM: 0.64; median Dice with TPM: 0.71; W = 80.5, p = 0.005). In contrast, adding a CRF did not enhance performance (median Dice without CRF: 0.75; median Dice with CRF: 0.72; W = 420.0, p = 0.69, N = 42), so we refrained from future use. This 3D CRF follows the structure of the MultiPrior model (28) implemented with a Gaussian filter and a bilateral filter, where we chose parameters for both filter widths (variance) to be 1, and the weight constant for both filters equal to 1, done with a simple grid search in the interval [0-5] on a small number of segmentations from the validation set. We then tested the U-Net with a TPM, where the TPM was a simple binary breast mask (Fig. 3). We found that the U-Net + TPM architecture outperformed the DeepMedic + TPM (Fig. S4A, Nt = 1025; Nv = 20; median Dice DeepMedic + TPM: 0.69; median Dice U-Net + TPM: 0.81; W = 26, p = 0.003). Next, we tested the U-Net without the use of the breast mask and found that performance was unchanged (Fig. S4B, Nt = 1025; Nv = 20; median Dice U-Net + Mask: 0.81; median Dice U-Net: 0.82; W = 113, p = 0.93). We tested whether addition of the T2 scans could improve performance, however, addition of coregistered T2 images actually decreased Dice score performance on the validation set (Fig. S4C, Nt = 2,555; Nv = 42; median Dice without T2: 0.75; median Dice with T2: 0.67; W = 188.0, p = 0.001). Finally, we compared the nonlinear intensity harmonization procedure (as described in Fig. S1) with a normalization that scales intensity separately for each exam. We found no difference in performance (Fig. S4D, Nt = 2,555; Nv = 42; median Dice nonlinear intensity harmonization: 0.80; median Dice exam-based scaling: 0.81; W = 515, p = 0.16), so we adopted the simpler scaling technique. Adding the slope of the contrast dynamic after the first contrast image significantly improved performance after training for 6 epochs in a balanced set containing equal number of malignant and benign breasts (Fig. S4E, Nt = 2,400; Nv = 100, median Dice without slopes: 0.60, median Dice with slopes: 0.66; W = 1836, p = 0.02).

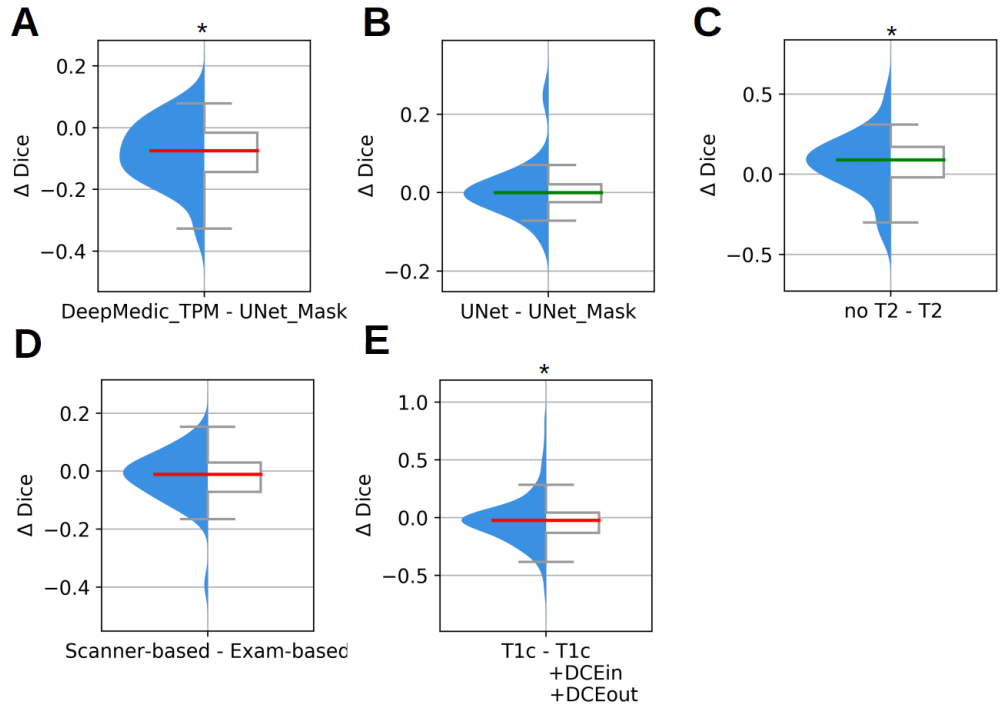

**Figure S4: Architecture optimization**. Difference in dice scores for pairwise comparison of five conditions for model selection. Medians of the difference shown as a red line if negative, and green if positive. **A:** Architecture based on the DeepMedic adding spatial priors against the U-Net. **B**: The U-Net against a U-Net using spatial information on breast tissue location. **C**: Addition or exclusion of T2-weighted scans. **D**: Two different methods for harmonization of voxel intensities in the images, namely nonlinear transformation dependent on the scanner manufacturer (see Fig. S1) against a simpler scaling method based on the T1-weighted scan of each exam. **E**: Benefit of addition of contrast enhancement dynamics (DCE-in, DCE-out). ( * : Wilcoxon p < 0.05).

## Training with Different Dataset Sizes

The basic hypothesis of this work is that training a deep network with a large dataset would significantly improve performance so as to reach radiologist-level performance. Performance gains with increased training set sizes are well established in machine learning. Nonetheless, we wanted to test this here explicitly for the typical dataset sizes used here and in the literature. When training this network on a typical dataset of 240 malignant and 240 benign breasts the network achieved a Dice score of 0.63 on a fixed validation set (N = 100). A tenfold larger training set with 2,400 malignant and 2,400 benign breasts achieved a Dice score of 0.69 in the same fixed validation set. Finally the validation Dice score on the full data set with two orders of magnitude more benign breasts (2,455 malignant and 60,108 benign) was 0.73. Both these gains in performance on a validation set were statistically significant (N = 100, W = 940, p = 0.001 small vs medium, N = 100, W = 1230, p = 0.03 medium vs large, see Fig. S5). Performance on this validation set (that is part of the training set) is lower than on the test set. The reason for this is that the test set contains more detailed segmentations necessary for multifocal cancers.

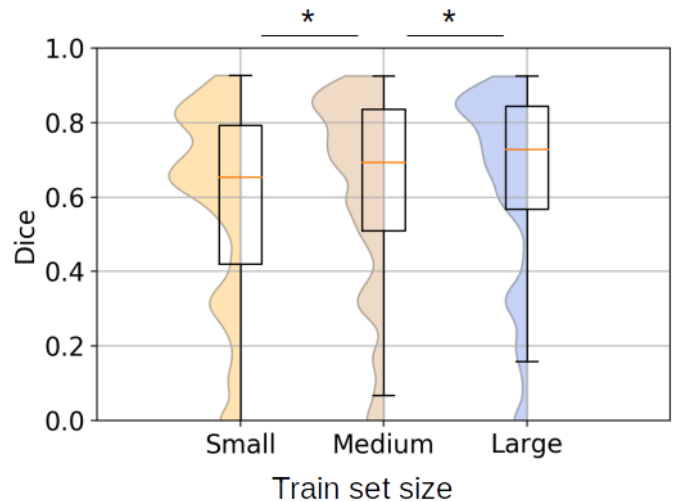

**Figure S5: Network performance as function of training set sizes**: Each of the sets has the same scans used for validation (100 malignant and 100 benign scans), varying only the amount for training: The number of malignant/benign scans available in each set are: 240/240 in the small set, 2,400/2,400 for the medium set, and 2,455/60,108 in the large (complete) set. The median Dice score is 0.66, 0.69 and 0.73 respectively (* indicates a pairwise significant difference between each group).

**Using a Consensus Segmentation as Reference**

To evaluate the performance on the test set, we constructed a reference segmentation by combining the segmentations of multiple radiologists. To prevent bias, this consensus segmentation did not include the segmentation of the radiologist being evaluated. Thus, each of the four radiologists had a separate reference segmentation that excluded the segmentation of the radiologist being tested. We used the intersection to combine the segmentations of the three other radiologists (AND), i.e. the three had a consensus on what voxels were part of the cancer. In a post hoc analysis we also considered the union of all voxels selected by any one of the radiologists (OR), as well as the majority vote (VOTE), i.e. voxels on which at least two radiologists agreed. The exact boundary of the cancer in the automated segmentation is dependent on the threshold applied to the output probabilities. To select the best threshold we computed the Dice score at different thresholds averaged over the four references (Fig. S6A) on 16 randomly selected cases for which we had segmentations from the four radiologists (Fig. 1). These 16 cases are separate from the 250 test cases and from the training or validation data (Fig. 1). The best threshold value in Figure S6A was 0.15 for the OR consensus, 0.35 for VOTE, and 0.60 for the AND consensus. Note that this threshold is relative to the maximum value in a test slice as described in the Methods section (Primary Outcome Measure). This relative threshold implements the prior knowledge (available to both the machine and the radiologist) that the 2D slice contains cancer.

Each type of reference segmentation (AND, OR, VOTE) differs in its degree of sensitivity and specificity, with AND being the most specific and OR the most sensitive. With the optimal threshold on the model's segmentations (Fig. S6B), we can see that radiologists 2 and 4 tend to be more specific, while radiologists 1 and 3 tend to be more conservative/sensitive (see example segmentations in Fig. S9B). For the AND and OR references, the network was equivalent in performance to the radiologists (i.e., the network was significantly better than the worst performing radiologists; OR median Dice R2: 0.71, median Dice machine: 0.80, W =

11633, p < .001; AND median Dice R1: 0.69, median Dice machine: 0.77, W = 10090, p < .001). For the VOTE reference, the network numerically outperformed the worst performing radiologist, but the difference was not statistically significant (median Dice R1: 0.79, median Dice machine: 0.81, W = 15282, p = 0.72).

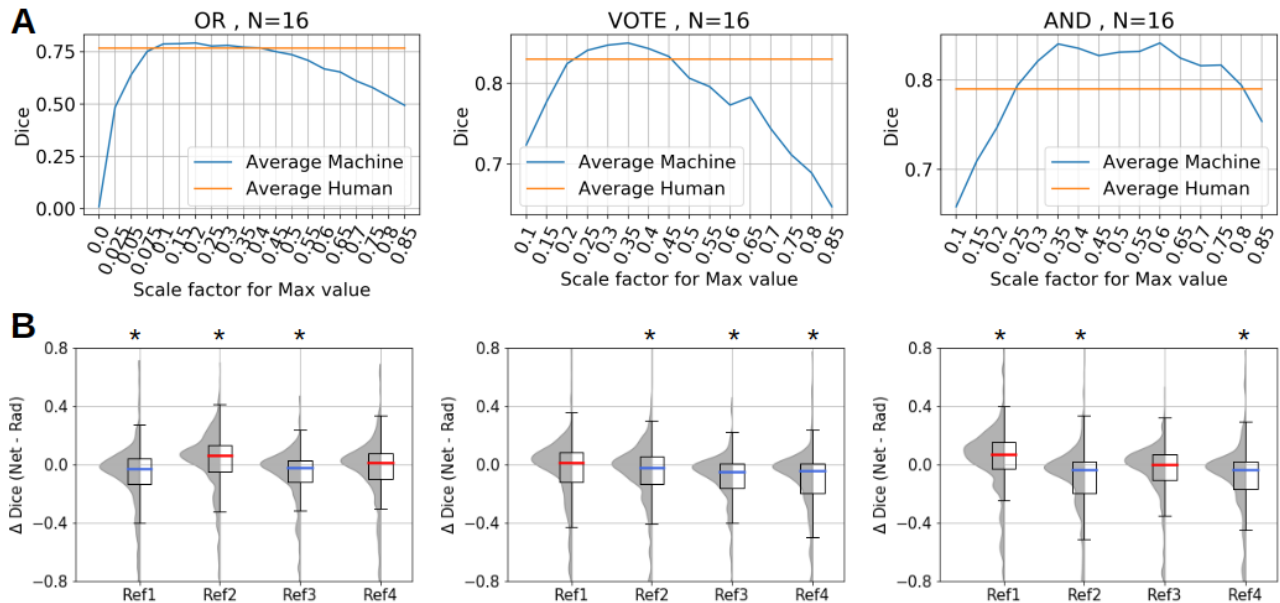

**Figure S6: Network performance as a function of threshold of the network output and comparison of different methods to obtain a consensus reference segmentation**. **A:** Dice score averaged over the four references and 16 exams for different threshold values over different consensus reference segmentations. **B:** Distribution across the 250 test cases of difference in Dice score between the network and each radiologist (Δ Dice) for each of the four reference segmentations for all three methods, respectively. The median Dice is either higher for the network (orange) or for the radiologist (blue). Significant difference of median from zero is indicated with * (Wilcoxon p < .001, Bonferroni multiple comparison correction).

### Evaluation of Criteria that Affect Segmentation Performance

Note that the presence of BPE affects network performance (Fig. S7B), but the effect is not strong. There are many cases with low BPE that also have poor Dice scores, for both the radiologists and the network (Fig. S8). Overall there is a significant correlation in performance between the network and radiologists (Spearman rank correlation between network and R1: r = 0.48, p < .001, network and R2: r = 0.39, p < .001, network and R3: r = 0.50, p < .001, network and R4: r = 0.42, p < .001). We also found a small effect of cancer size on Dice score performance (Fig. S7C, Spearman rank r = 0.19, p = 3e-3, N = 250), which is to be expected, given the definition of Dice score. Note that when the machine has poor Dice scores, the radiologists may still perform well (points on the top left corner of scatter plots in Fig. S8 and S9). This is necessarily true due to the definition of the consensus references.

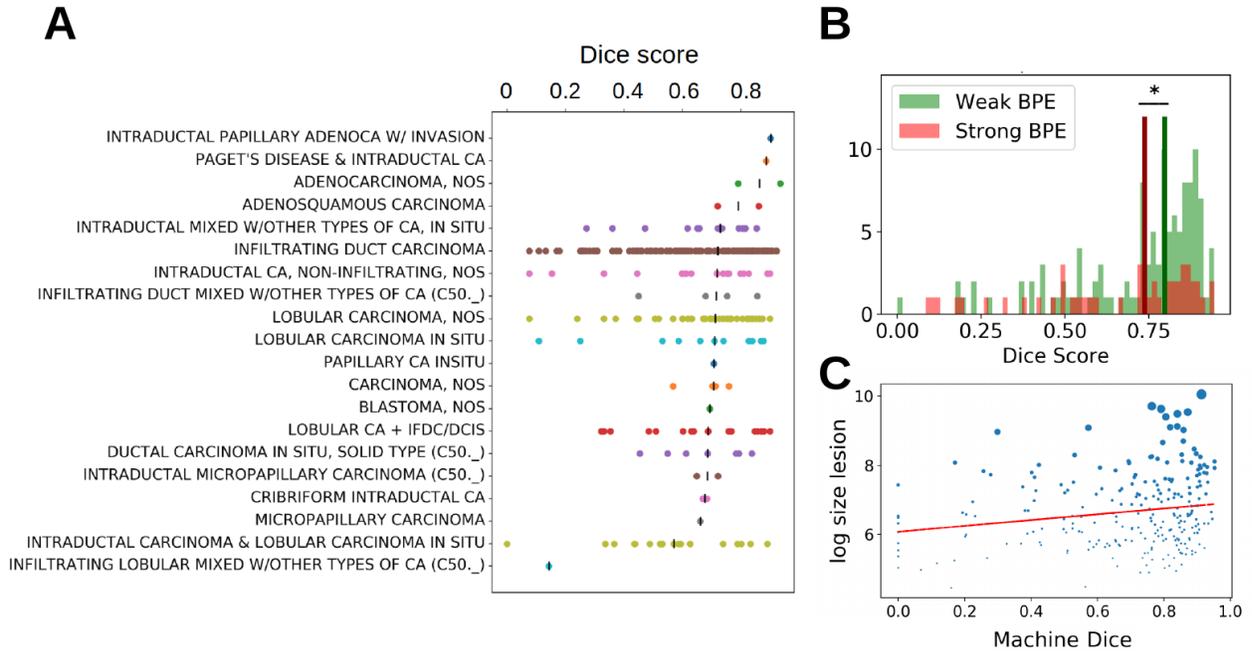

**Figure S7: Segmentation performance of the network depends on BPE and type of tumor.**
**A:** For each tumor type, the performance of the model is displayed, and the median is marked with a black line (ANOVA, F = 2.12, p = 0.004, df = 262). **B**: Histograms showing Dice scores for the network, depending on the strength of BPE as assessed in the radiology reports. Radiologist performance is not affected by breast density (not shown), whereas the model's performance decays in the presence of strong BPE (median Dice $\leq$ 75% BPE: 0.79; median Dice $\geq$ 75% BPE: 0.73; U = 2592; p = 0.02, Mann Whitney-U). **C:** Correlation of model performance and the size of the cancer in each scan. Each dot corresponds to a scan and the size is directly proportional to the cancer area.

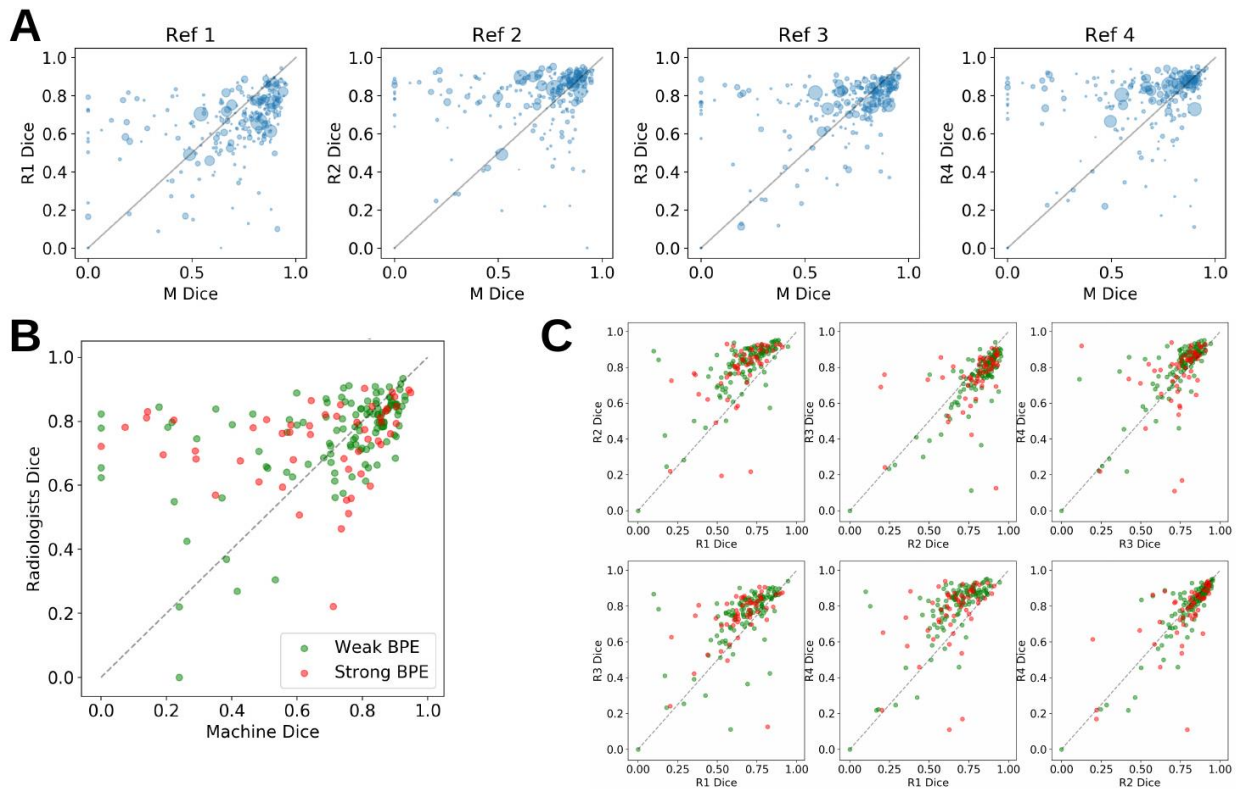

**Figure S8: Correlation of radiologist and network performance. A:** Correlation of performance in Dice score across all 250 scans in the test set, between each radiologist (R1-R4 Dice) and the network (M Dice). Each dot corresponds to a scan and its size is proportional to the cancer area. **B:** Correlation of average radiologist and network performance, color coded by weak BPE (green) and strong BPE (red) for 173 scans which had explicit description of BPE. **C:** Correlation of performance between all 4 radiologists, color coded by BPE strength.

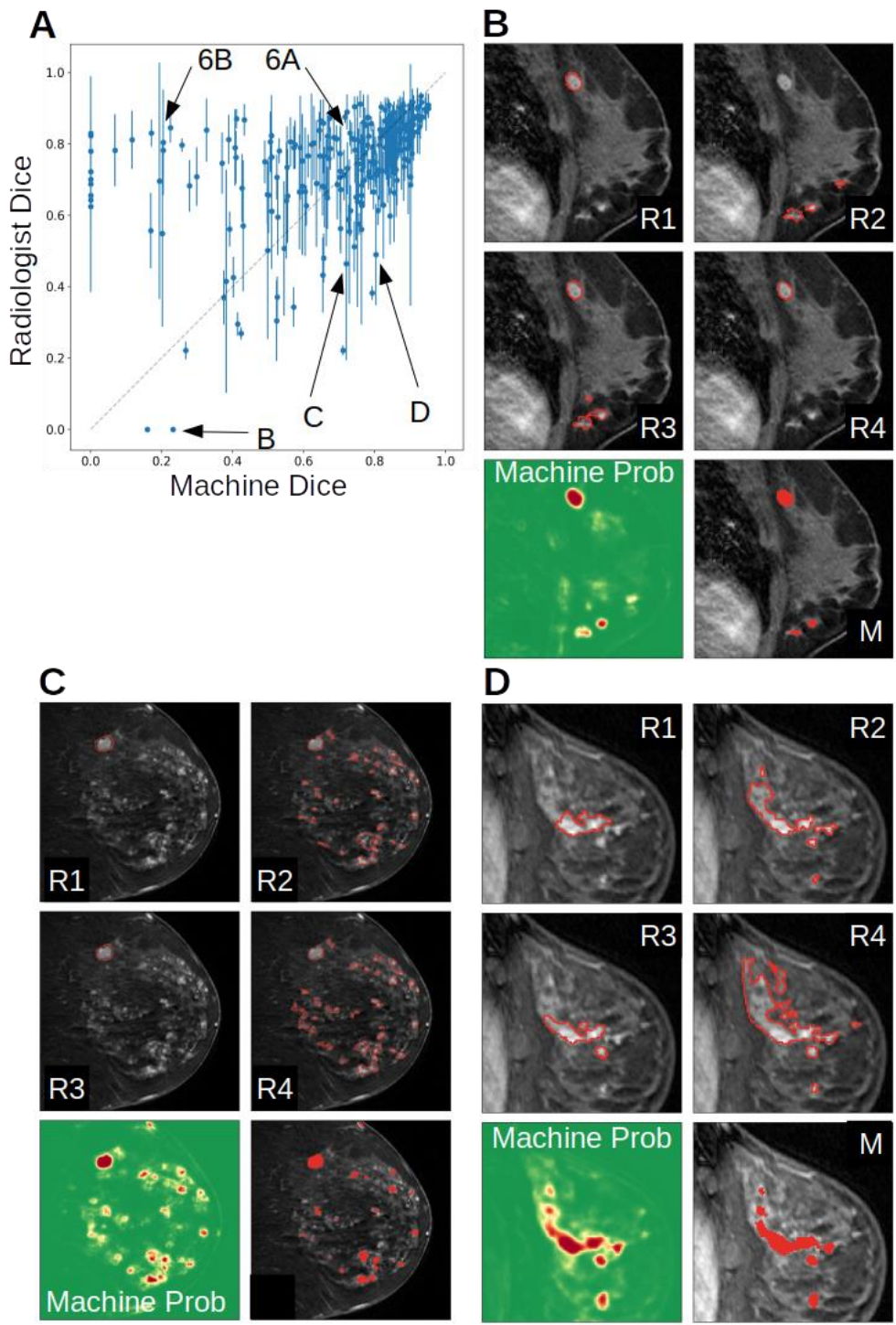

**Figure S9: Agreement between radiologists and network. A:** Correlation scatter plot between averaged radiologists (Radiologist Dice) and the network (Machine Dice) using as reference the consensus intersection (AND). Error bars show the standard deviation across all four radiologists. Three scans are selected for detailed viewing on panels B,C and D. Scans marked as 6A and 6B are displayed respectively on Figure 6 of the main text. **B:** Example of one of the scans with zero radiologist performance, where there is strong disagreement between radiologists and the network finds all common locations with different degrees of certainty. After thresholding, the network agrees with R1 and R4. **C:** Two pairs of radiologists agree on the level of sensitivity to segment locations: R1 with R3 and R2 with R4. **D:** Radiologists agree on the main location of the cancer but disagree on the boundaries.

## Comparison with Fuzzy-c-means

We implemented a conventional image segmentation algorithm (fuzzy-c-means, (35)) to test the performance of this dataset and provide a baseline. This was implemented in Python with the package (skfuzzy), and images were pre-processed to increase contrast (using skfuzzy.contrast). This is a recommended preprocessing step for this method and increased performance of the algorithm on training data. The same data were used as for the network (T1c, DCE-in, DCE-out). Parameters of the algorithm were optimized on a validation set of 30 malignant breasts taken from the training set. The Dice score evaluated on the consensus reference (AND) for the test set show that the method substantially underperforms the network (Fig S10; median Dice score for network vs fuzzy-c-means: 0.77 vs 0.10, Wilcoxon $W = 475$, $p < .001$, $N = 250$). This highlights the challenging nature of this task.

Note that in previous work with fuzzy-c-means, segmentation of breast lesions was evaluated within a manually-selected rectangular region of interest (ROI) (26), i.e. the segmentation was only semi-automatic. When we limit the evaluation here to a rectangular bounding box around the reference segmentation (extending an extra 25% on all sides), performance of the fuzzy-c-means improves to 0.65 and to 0.82 for the network (ES = 0.82, $W = 2927$, $p < .001$, $N = 250$).

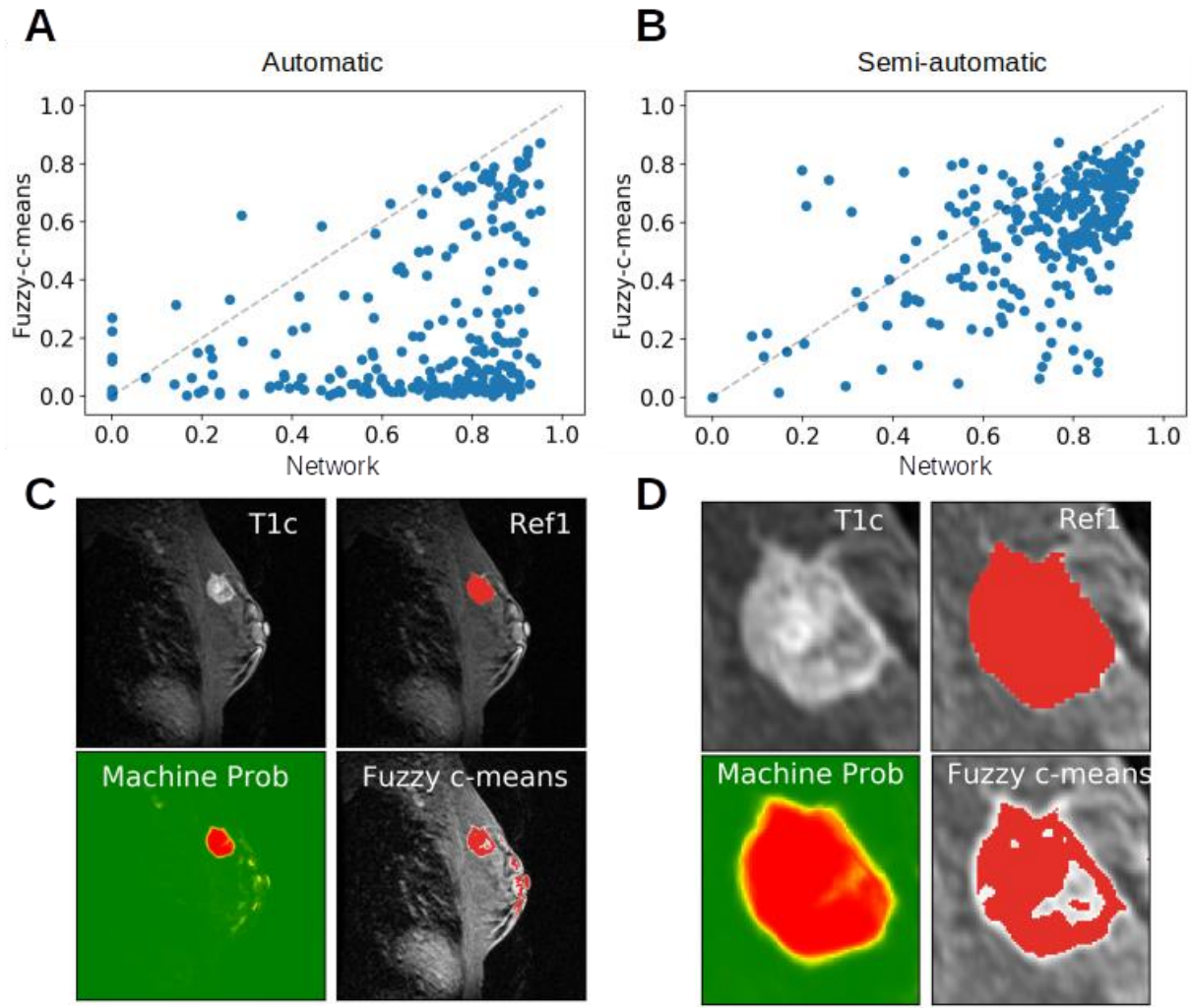

**Figure S10: Segmentation performance of a conventional image segmentation algorithm. A:** Dice scores comparing the network vs a fuzzy c-means algorithm on the entire 2D slice (Automatic). Each point is one of 250 test cases. **B:** Same comparison as in A, but now limiting evaluation to manually preselected ROI (semi-automatic; for example see Panel D). **C:** Example of a breast scan (same data as used for the network, DCE-in and DCE-out not shown), a reference segmentation (Ref1), the probabilistic output of the network (Machine Prob), and segmenting with fuzzy-c-means. **D:** Region of interest for semi-automatic evaluation for the same scan as in panel C.

**Pairwise Comparison of Radiologists with Network**

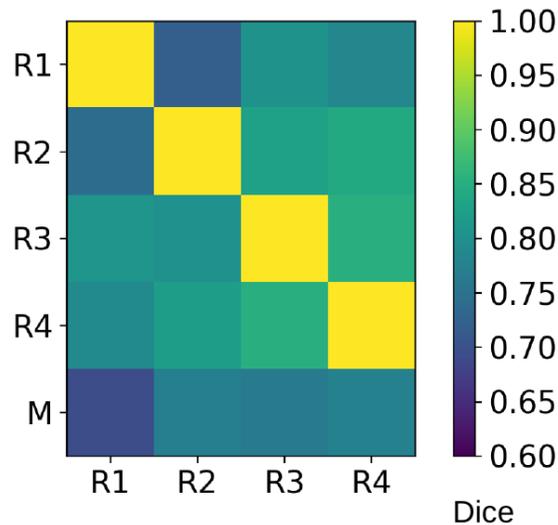

Figure S11: Dice score (color coded) quantifying agreement between each radiologist's manual segmentation and that of another radiologist as well as the machine's segmentation (mean across 250 test cases).

**Comparison of Dice scores using repeated measures ANOVA**

To test for differences between radiologists and the machine in Dice score, while taking into account repeated cases we used a repeated measures ANOVA with cases as random effect (N = 250) and fixed effects of reference segmentations (Ref1, Ref2, Ref3, Ref4) and segmenter (either radiologist or machine). There was a strong difference in Dice score across cases ($F(249) = 29.67$, $p < .001$), which is evident already in the long tails of Fig 5A. There was no significant difference across references ($F(3) = 1.02$, $p = 0.49$) and no difference between the machine and the radiologist ($F(1) = 1.76$, $p = 0.28$) suggesting that on average, the machine has comparable performance to the radiologists. There was an interaction between reference and segmenter ($F(3) = 78.42$, $p < .001$) suggesting that the machine may have outperformed some of the radiologists, while other radiologists may have outperformed the machine, which is evident in Fig 5B. There was also a strong interaction between case and segmenter ($F(249) = 14.11$, $p < .001$) which means that some cases were better segmented by the radiologists, whereas others were better segmented by the machine. Similar results were observed when performing this ANOVA with logit-transformed Dice scores.

Given that neither the Dice scores nor logit-transformed Dice scores were normally distributed, in the main results section we used the Wilcoxon signed-rank test as a non-parametric test. That analysis focused on the mean difference between radiologist and machine (fixed effect of segmenter, and paired to account for repeated measures). In follow-up comparisons we tested for the difference between radiologist and machine for each reference (again paired), which confirmed the interaction observed here between segmenter and reference. These individual follow-up tests, which include comparisons with the worst and best performing radiologist, were also an integral part of the planned TOST procedure.

| Cancer type | Training | Validation | Test |
| --- | --- | --- | --- |
| INFILTRATING DUCT CARCINOMA | 1781 | 61 | 116 |
| LOBULAR CARCINOMA IN SITU | 381 | 7 | 10 |
| LOBULAR CARCINOMA, NOS | 325 | 9 | 33 |
| LOBULAR CA + IFDC/DCIS | 284 | 9 | 14 |
| INTRADUCTAL CA, NON-INFILTRATING, NOS | 175 | 4 | 13 |
| INTRADUCTAL MIXED W/OTHER TYPES OF CA, IN SITU | 149 | 3 | 12 |
| INTRADUCTAL CARCINOMA & LOBULAR CARCINOMA IN SITU | 115 | 4 | 16 |
| INFILTRATING DUCT MIXED W/OTHER TYPES OF CA (C50._) | 44 | | 3 |
| DUCTAL CARCINOMA IN SITU, SOLID TYPE (C50._) | 36 | | 7 |
| CRIBRIFORM INTRADUCTAL CA | 24 | 2 | 2 |
| CARCINOMA, NOS | 19 | 1 | 5 |
| ADENOCARCINOMA, NOS | 18 | 3 | 2 |
| INTRADUCTAL MICROPAPILLARY CARCINOMA (C50._) | 13 | 2 | 1 |
| TUBULAR ADENOCARCINOMA | 12 | | |
| PAGET'S DISEASE & INTRADUCTAL CA | 11 | | 1 |
| METAPLASTIC CARCINOMA, NOS | 11 | | |
| MUCINOUS ADENOCARCINOMA | 10 | | |
| NON-INFILT INTRADUCTAL PAPILL ADENOCA | 9 | 1 | |
| PAGET'S DISEASE & INFILT DUCT CA | 8 | | |
| NO CLM ENTRY | 7 | | |
| INTRADUCTAL PAPILLARY ADENOCA W/ INVASION | 6 | | 1 |
| INFILTRATING LOBULAR MIXED W/OTHER TYPES OF CA (C50._) | 4 | 1 | 1 |
| MICROPAPILLARY CARCINOMA | 4 | | 1 |
| APOCRINE ADENOCARCINOMA | 4 | | |
| MAL LYMPHOMA, LARGE CELL, DIFFUSE | 4 | | |
| ADENOID CYSTIC CARCINOMA | 3 | | |
| PAGET'S DISEASE, MAMMARY | 3 | | |
| PAPILLARY CARCINOMA, NOS | 3 | | |
| BLASTOMA, NOS | 2 | | 1 |
| HEMANGIOSARCOMA/ANGIOSARCOMA | 2 | | |
| MARGINAL ZONE B-CELL LYMPHOMA, NOS | 2 | | |
| PAPILLARY CA INSITU | 1 | | 1 |
| ADENOCA W/ SQUAMOUS METAPLASIA | 1 | | |
| GIANT CELL SARCOMA | 1 | | |
| INFLAMMATORY CARCINOMA | 1 | | |
| LEIOMYOSARCOMA, NOS | 1 | | |
| MYXOID LIPOSARCOMA | 1 | | |
| PHYLLODES TUMOR MALIGNANT | 1 | | |
| PLEOMORPHIC CARCINOMA | 1 | | |
| ADENOSQUAMOUS CARCINOMA | | | 1 |

**Table S2:** Cancer types in the training, validation and test sets.

| Age interval | Training set counts | Validation set counts | Test set counts |
|---|---|---|---|
| 10-20 | 24 | 0 | 0 |
| 20-30 | 885 | 2 | 2 |
| 30-40 | 5649 | 23 | 25 |
| 40-50 | 17842 | 54 | 71 |
| 50-60 | 19398 | 55 | 67 |
| 60-70 | 9639 | 36 | 41 |
| 70-80 | 2621 | 9 | 24 |
| 80-90 | 183 | 2 | 1 |
| 90-100 | 4 | 0 | 0 |

**Table S3:** Ages of patients at the available exams in the training, validation and test sets.